\documentclass[a4paper,fleqn]{cas-dc}

\usepackage[numbers]{natbib}
\usepackage{multirow}
\usepackage{amsmath}
\usepackage{amssymb}
\usepackage{amsfonts}
\usepackage{graphicx}
\usepackage{bm} 
\usepackage{graphicx}
\usepackage{subcaption}
\usepackage{booktabs}
\usepackage{caption} 

\usepackage[accsupp]{axessibility} 

\ExplSyntaxOn
\cs_set:Npn \__first_footerline:
{
  \group_begin:
  \small
  \sffamily
  \ifnum\theblind>0\relax
  \else
    \__short_authors: :~
  \fi
  { \rmfamily \itshape The~paper~is~under~consideration~at~Pattern~Recognition~Letters }
  \group_end:
}
\ExplSyntaxOff

\def\tsc#1{\csdef{#1}{\textsc{\lowercase{#1}}\xspace}}
\tsc{WGM}
\tsc{QE}
\tsc{EP}
\tsc{PMS}
\tsc{BEC}
\tsc{DE}

\begin{document}
\let\WriteBookmarks\relax
\def\floatpagepagefraction{1}
\def\textpagefraction{.001}
\shorttitle{Field-Localized Forgery Detection}
\shortauthors{Abhishek et~al.}

\title [mode = title]{ Field-Localized Forgery Detection for Digital Identity Documents}                      



\author[1]{Abhishek Kumar}
\cormark[1]
\ead{akumar@turing.ac.uk}
\credit{Conceptualization, Methodology, Software, Writing -- original draft}

\affiliation[1]{organization={The Alan Turing Institute},
                city={London},
                country={United Kingdom}}

\author[2]{Riya Tapwal}
\ead{riya@iitmandi.ac.in}
\credit{Methodology, Software, Investigation}

\affiliation[2]{organization={Indian Institute of Technology Mandi},
                city={Mandi},
                state={Himachal Pradesh},
                country={India}}

\author[1]{Carsten Maple}
\ead{cmaple@turing.ac.uk}
\credit{Supervision, Writing -- review \& editing}

\author[1]{Mark Hooper}
\ead{mhooper@turing.ac.uk}
\credit{Supervision, Writing -- review \& editing}

\cortext[1]{Corresponding author}

\begin{abstract}
Digital onboarding and eKYC systems used by banks, fintech platforms, telecom providers, and other third-party services commonly verify users by comparing an uploaded identity document with a selfie or live facial capture. This workflow is convenient, but it also makes verification systems vulnerable to localised document manipulations, such as replacing the facial photograph, editing textual identity fields, or altering both. Existing image-forgery detectors are largely designed for natural images and do not explicitly account for the structured layout of identity documents, where security-relevant information is concentrated in specific semantic fields. We propose FLiD, a lightweight field-localised framework for identity-document forgery detection. Instead of processing the full document image, FLiD localises the facial and textual regions using a fine-tuned YOLO11 detector, extracts compact representations with a frozen MobileNetV3-Small backbone, and classifies field-level forgeries using a small 191K-parameter head. Face and text detectors are trained independently and combined through score-level fusion for documents containing simultaneous manipulations. Under 5-fold cross-validation, FLiD achieves AUC scores of 0.834, 0.926, and 0.837 for face, text, and combined attacks, respectively, while reducing Equal Error Rate by 28--29 percentage points compared with a full-document baseline. FLiD also outperforms general-purpose forgery detectors while requiring 13$\times$ fewer trainable parameters and 21$\times$ fewer FLOPs per field ($\sim\!8\times$ per full multi-field document), making it a promising candidate for data and resource constrained KYC deployment.
\end{abstract}




\maketitle

\section{Introduction}
\label{sec:intro}

Forgery detection in identity documents has become a critical challenge in modern digital identity verification systems, where banks, fintech platforms, telecom providers, and other third-party services routinely require users to upload photographs or scanned copies of identity documents for remote authentication~\cite{laas2022promises,eba2022remote}. The reliance on such remote onboarding workflows increased significantly following the COVID-19 pandemic~\cite{laas2022promises}, when physical verification and contact-based biometric modalities such as fingerprint acquisition were substantially replaced by fully digital verification pipelines~\cite{zhang2025distilled}. Consequently, identity verification systems increasingly rely on visual inspection of submitted document images rather than direct access to centralised identity databases~\cite{shi2019docface+, albiero2020identity}. These systems often combine document analysis with face matching between the identity card and a live selfie. This dependence on document images makes them vulnerable to localised manipulations, face replacement, text editing, or combined tampering, which can enable identity fraud while remaining visually difficult to detect through manual inspection~\cite{tapia2024first, zheng2019survey}. The scale of this threat has increased substantially: recent identity-verification data show that digitally forged identity documents accounted for 57.46\% of all detected document fraud in 2024, compared with 16.7\% in the previous year.  National identity cards were the most frequently targeted document category, accounting for 40.8\% of document-fraud attempts~\cite{entrust2025identityfraud}.


Identity documents differ fundamentally from natural images: they contain
structured layouts composed of semantically meaningful fields such as
facial photographs, textual identity information, security textures, and
machine-readable zones~\cite{shi2019docface+, tapia2024first}. Even small
localised modifications, such as altering a date of birth or swapping the
facial photograph, can compromise system integrity and lead to financial
or regulatory risks~\cite{zheng2019survey}. Figure~\ref{fig:sample} illustrates a representative combined forgery, where both the facial photograph and multiple textual identity fields have been manipulated and highlighted. In our evaluation, we consider three attack scenarios: face-only manipulation, text-only manipulation, and combined face--text manipulation. The example shows how localised edits to semantically important fields can remain visually subtle while compromising document authenticity.

\begin{center}
\begin{minipage}{\columnwidth}
\centering
\includegraphics[width=0.97\columnwidth]{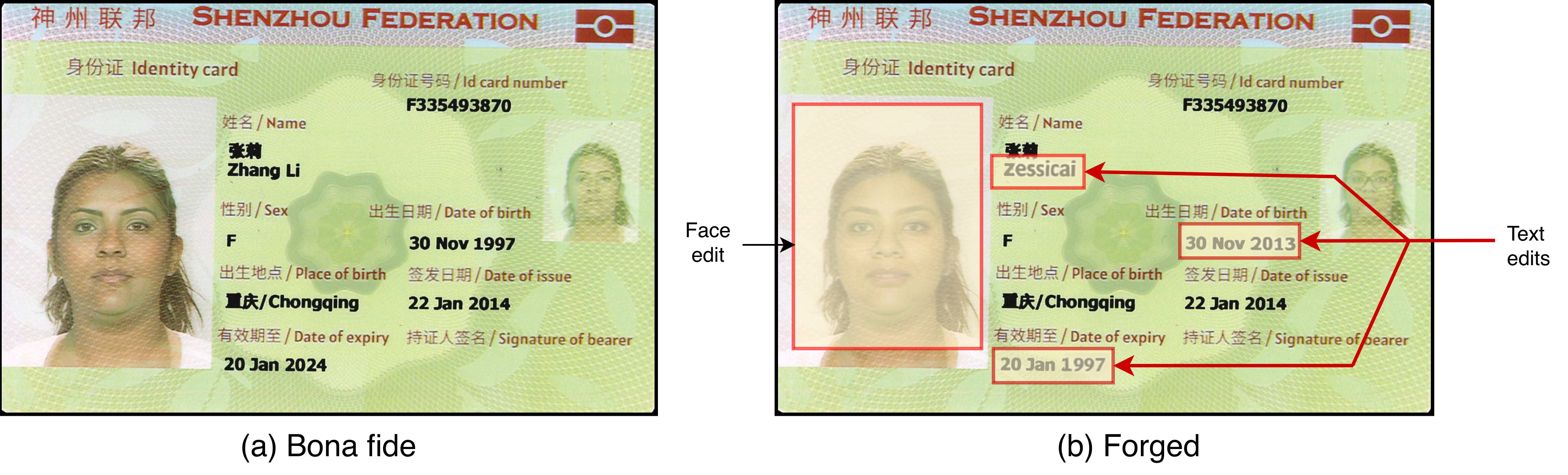}
\captionof{figure}{Example of a combined forgery with changes to the face and text fields. Edited regions are highlighted.}
\label{fig:sample}
\end{minipage}
\end{center}

Only a few public studies have directly examined identity-document forgery detection, as privacy, regulatory, and security constraints keep most industrial datasets and deployed pipelines proprietary, limiting reproducibility and motivating open, efficient, domain-specific approaches~\cite{pad2025pattern, occludedid2023}. In parallel, general image forgery detection has been extensively studied using methods such as TruFor~\cite{guillaro2023trufor}, UniVAD~\cite{yang2025univad}, and MMFusion~\cite{mmfusion2024}, which achieve strong performance on natural-image manipulations through transformer-based architectures, multimodal fusion, and anomaly-based localisation. However, these methods are designed and evaluated on natural-image benchmarks, where the goal is to detect object insertion, removal, or generative synthesis; they are not optimised for the structured, multimodal characteristics of identity documents, and their accuracy degrades to near chance when applied directly to structured identity data\cite{korshunov2025fantasyid}.

To address these challenges we propose \textbf{FLiD} (Field-Localized Identity Document forgery detection), a lightweight, field-aware framework that explicitly models identity documents as structured collections of semantic fields. Rather than processing the entire document indiscriminately, FLiD focuses on the regions most susceptible to tampering, the facial photograph and the textual identity fields. A fine-tuned object detector first localises and extracts these regions, enabling consistent preprocessing across heterogeneous layouts. The regions are encoded into compact embeddings by a \emph{frozen} ImageNet-pretrained backbone, and a lightweight classifier performs field-level forgery detection. For documents with simultaneous manipulations, the per-field detectors are combined by score-level fusion, flagging a document if any field appears forged.

Under $5$-fold cross-validation, FLiD attains AUC scores of $0.834$ (face),
$0.926$ (text), and $0.837$ (combined), whereas a full-document baseline
trained from scratch~\cite{pad2025pattern} collapses to near-chance
discrimination ($0.47$--$0.53$ AUC) under the identical protocol. With
manually annotated ground-truth crops as an upper bound, AUC reaches
near-perfect for text ($0.998$), showing that more precise field
localisation translates directly into higher detection accuracy. This
indicates that field localisation not only improves forgery-detection
performance but also enables the model to learn tampering cues that
generalise across documents while requiring substantially less computation,
suiting FLiD for resource-constrained KYC deployment.

\paragraph{\textbf{Contributions.}}
\begin{itemize}
\item \textbf{Field-localised detection framework.} We introduce FLiD, a
forgery-detection framework for structured identity documents that models
each document as a composition of semantic fields and analyses the face and
text regions individually, rather than treating the document as an
undifferentiated image.

\item \textbf{Data and compute-efficient design.} FLiD uses a frozen ImageNet-pretrained MobileNetV3-Small backbone and a compact $191$K-parameter classification head, achieving strong performance with $13\times$ fewer trainable parameters and up to $21\times$ fewer FLOPs per field ($\sim!8\times$ per full multi-field document) than the full-document baseline.

\item \textbf{Cross-attack evaluation and localisation ablation.} We evaluate FLiD under identity-disjoint 5-fold cross-validation and a cross-attack setting in which face and text detectors trained on single-field attacks are fused at inference for unseen combined attacks. We further quantify the effect of ground-truth, automatic and coarse crop strategies on detection performance.

\end{itemize}

\section{Related Work}
\label{sec:related}

\textbf{Identity-document forgery and PAD.} Identity-focused work emphasizes mobile capture conditions and presentation attacks. Studies on physically occluded fake documents \cite{occludedid2023} and ID-card PAD in remote verification systems \cite{pad2025pattern} highlighted print–scan and replay attacks, but typically reused generic architectures. PSFNet \cite{psfnet2022} targeted passport security features, while texture-based analyses \cite{texture2025spie} examined print–scan inconsistencies. The Document Liveness Challenge dataset \cite{dlc2022} enabled benchmarking under recapture conditions. Further, ICDAR competitions on tampered text \cite{icdar2023tamper} and receipt forgery \cite{icdar2023receipt} encouraged document-centric evaluation. Learning-based pipelines for smartphone IDs \cite{mlframeworkid2023} attempted unified modeling but rarely captured cross-field dependencies explicitly. Shi and Jain~\cite{shi2019docface+} proposed \textit{DocFace}, a deep learning framework for matching ID document photographs with live selfie images using partially shared networks and dynamic weight imprinting for shallow datasets. The importance of presentation attack detection has also been highlighted through benchmarking efforts such as the PAD-IDCard competition~\cite{tapia2024first}. Beyond document-specific studies, Zheng et al.~\cite{zheng2019survey} provide a survey of digital image tampering detection methods. Related work has also examined challenges in ID-to-selfie matching~\cite{albiero2020identity} and face-based active authentication on mobile devices~\cite{perera2018face,fathy2015face}, highlighting issues such as illumination variation, motion blur, and pose changes in mobile capture conditions.

\textbf{Generic Image Forgery Detection.} Early modern learning-based detectors focused on low-level inconsistencies across entire images. MVSS-Net \cite{chen2021mvss} demonstrated that multi-view and multi-scale supervision improves localization, but incurred high computational cost and remained agnostic to semantics. CAT-Net \cite{kwon2021catnet} traced compression artifacts left by splicing operations, but was limited to JPEG-style pipelines. PSCC-Net \cite{liu2022pscc} refined masks using progressive spatial–channel modeling, yet still relied on dense feature maps. Later, TruFor \cite{guillaro2023trufor} unified RGB and trace features to predict manipulation masks and reliability maps, while HiFi-Net \cite{guo2023hifi} introduced hierarchical supervision over forgery attributes. DiffForensics \cite{yu2024diffforensics} incorporated diffusion priors to improve generalization, and FatFormer \cite{liu2024fatformer} adapted transformers for synthetic imagery. Despite improved robustness, these methods treat images as homogeneous grids and explain decisions through pixel heatmaps. Further, UniVAD \cite{yang2025univad} performs training-free anomaly detection by identifying deviations from normal data, but it does not explicitly model intentional, field-specific edits common in identity fraud.

\textbf{Structured and Field-Level Forgery.} Document-specific systems such as DCLNet \cite{dclnet2024} and edge-focused detectors \cite{edge2024} exploited boundaries and region cues. ID-centric PAD systems \cite{pad2025pattern,occludedid2023} and passport verification networks \cite{psfnet2022} similarly focused on localized components. However, MMFusion \cite{mmfusion2024} fused heterogeneous cues but still operated over dense spatial grids. Existing systems thus encode document structure only implicitly, motivating our explicit field-token representation and cross-field reasoning.

\section{Methodology}
\label{sec:method}
We propose a field-localised forgery detection system that targets the semantically meaningful regions of a digital identity document rather than the document as a whole. Rather than learning dense pixel-level representations as for natural images, we extract compact features from the fields most susceptible to tampering, field-level extraction, frozen pretrained features, and a lightweight per-field classifier, yielding reliable, reproducible, and edge-deployable detection.

\begin{figure*}[t]
    \centering
    \includegraphics[width=0.75\linewidth]{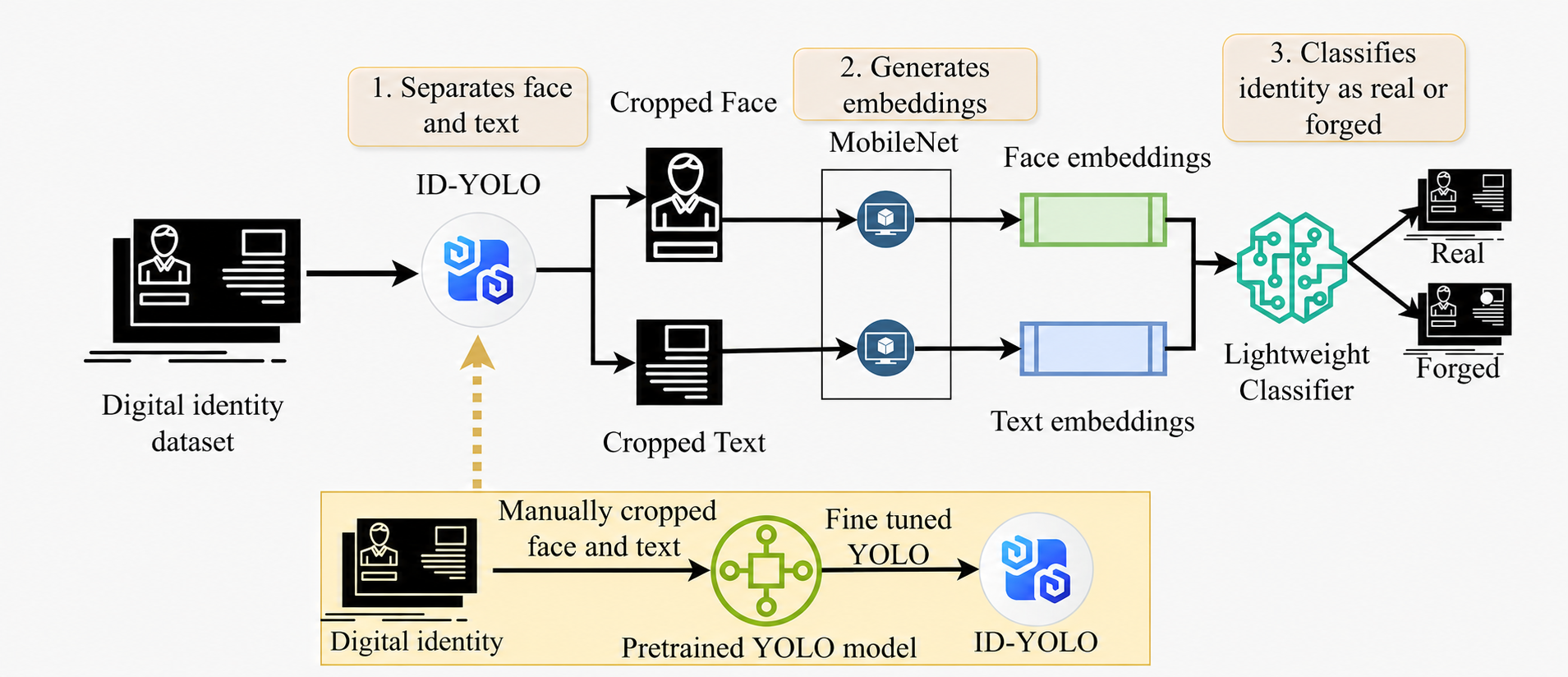}
    \caption{Overview of FLiD. The system first localises semantically meaningful identity fields, then encodes face and text crops independently using a frozen MobileNetV3-Small backbone. Lightweight field-level classifiers produce bona-fide/forged scores, which are fused at document level for combined attacks.}
    \label{fig:methodology}
\end{figure*}

As shown in Fig.~\ref{fig:methodology}, the pipeline has three stages: (i)~a fine-tuned YOLO11 detector localises the facial photograph and textual sub-fields, obtained by fine-tuning a pretrained YOLO11 on manual face/text annotations; (ii)~each cropped field is encoded independently by a frozen MobileNetV3-Small backbone into a 576-dimensional embedding; and (iii)~a lightweight classifier maps each embedding to a bona-fide/forged decision, with per-field decisions for multi-field documents combined via score-level fusion (Section~\ref{sec:fusion}) rather than a jointly trained fusion model, so combined-attack documents are never needed during training.

\subsection{Identity Field Localisation with YOLO11}
\label{sec:yolo}
Identity documents vary substantially in layout and field placement across issuing authorities; FantasyID alone spans $13$ countries. We first manually annotated and cropped the face and text regions for all images, producing paired full-document and cropped-field versions that (i) enable direct comparison between full-document and field-level analysis and (ii) provide training labels for an automatic field localiser. We adopt YOLO11 for its stable implementation, strong detection performance, and availability of pretrained weights; as localisation is only a preprocessing step, it adds no
architectural complexity to the classification pipeline.

Given an input document $I$, the detector predicts boxes
$\mathcal{B}=\{b_1,\dots,b_K\}$ with $b_i=(x_i,y_i,w_i,h_i,c_i)$
(centre, size, and class
$c_i\in\{\text{face},\text{name},\text{dob},\text{doe},\text{doi}\}$),
and each field is extracted by cropping $I_i=\mathrm{Crop}(I,b_i)$.
The pretrained detector localised fields inconsistently across heterogeneous
layouts, so we fine-tuned it on the manual annotations; the fine-tuned
model then reliably extracts face and text regions across document formats.
All detected text-field crops $\{I_j : c_j \in \{\text{name},\text{dob},\text{doe},\text{doi}\}\}$ are embedded and scored individually, while the largest detected face crop provides the face embedding $\mathbf{e}_f$; the per-field scores are aggregated only at the decision stage.

\paragraph{Crop strategies and evaluation tiers.}
To decouple the effect of field localisation quality from classification
quality, we evaluate three crop sources. \emph{Ground-truth (GT) crops}
are obtained from manual annotations and serve as an upper bound on
detection accuracy; they require human effort and are not available at
deployment time. \emph{YOLO11 crops} are produced fully automatically by
the fine-tuned detector and represent our deployable pipeline, all main
results are reported under this setting. \emph{Coarse crops} use a single
bounding box covering the entire text area without field-level precision,
serving as a lower bound that requires only coarse document segmentation.
The gap between GT and YOLO11 reflects the room for improvement available
from better automatic field localisation.

\subsection{Field-Level Feature Encoding}
\label{sec:encode}
Each detected field is encoded by a frozen MobileNetV3-Small backbone
pretrained on ImageNet-1K, chosen for its accuracy/efficiency trade-off
under the latency constraints of large-scale verification. The backbone
weights remain \emph{frozen}; only the downstream classifier is trained.
This avoids overfitting to the small identity-document corpus while
retaining rich pretrained features, an effect we quantify in
Section~\ref{sec:results}, where training a full-document classifier from
scratch collapses to near-chance.

Let $I_f$ be the cropped face and ${I_j}_{j\in\mathcal{T}}$ the cropped text sub-fields. Each is resized to $224\times224$, normalised with ImageNet statistics, and passed through the encoder $\phi(\cdot)$:

\[
\mathbf{e}_f=\phi(I_f),\qquad \mathbf{e}_t^{(j)}=\phi(I_j)\ \ \text{for } j\in\mathcal{T},\qquad \mathbf{e}_f,\mathbf{e}_t^{(j)}\in\mathbb{R}^{576},
\]

where $\mathcal{T}$ indexes the detected text sub-fields and embeddings
are taken from the final global-average-pooling layer. These capture
identity-relevant cues such as skin-texture consistency, character-rendering
artifacts, and localised editing inconsistencies. Embeddings are
precomputed once and stored, so classification operates directly on the
$576$-D vectors. This reduces the trainable parameter count to $191$K,
versus $2.55$M for the from-scratch full-document baseline.

\subsection{Field-Level Classification and Cross-Attack Score Fusion}
\label{sec:fusion}
A single-field embedding $\mathbf{e}\in\mathbb{R}^{576}$ ($\mathbf{e}_f$ or
$\mathbf{e}_t$) is classified by a lightweight feed-forward head
$g(\cdot)$ of $L{=}5$ fully connected layers with ReLU and dropout,
trained with class-balanced binary cross-entropy
($w^{+}=n_{\text{neg}}/n_{\text{pos}}$):
\[
\mathcal{L}=-\bigl[w^{+}y\log\sigma(\hat{y})
+(1-y)\log\bigl(1-\sigma(\hat{y})\bigr)\bigr],
\]
where $\hat{y}$ is the output logit, $y\in\{0,1\}$ the bona-fide/forged
label, and $\sigma(\cdot)$ the sigmoid. We train two independent detectors, $g_f$ (face) and $g_t$ (text). The face
detector scores the face crop, $P_f(\text{real})=\sigma(g_f(\mathbf{e}_f))$, and
the text detector scores each text field, $P_t^{(j)}(\text{real})=\sigma(g_t(\mathbf{e}_t^{(j)}))$.

\paragraph{Combined attacks.} A document is authentic only if every field is authentic, so we score each field independently and let the most suspicious field decide:
\[
P_{\text{doc}}(\text{real})=\min\!\Bigl(P_{\!f}(\text{real}),\;\min_{j\in\mathcal{T}}P_{\!t}^{(j)}(\text{real})\Bigr),
\]
This prevents a single tampered field from being diluted by genuine fields. The face and text detectors are trained only on single-field data; combined-attack documents are used exclusively at test time (Section~4.1), reflecting realistic deployment and avoiding overfitting to combined-attack samples.

\section{Performance Evaluation}
\label{sec:results}

\subsection{Experimental Setup}
\label{sec:setup}

All experiments use a two-stage pipeline. Region-of-interest (ROI) fields are first localised with a fine-tuned YOLO11 detector, after which $576$-dimensional embeddings are extracted from each region using a MobileNetV3-Small backbone pretrained on ImageNet-1K with its classification head removed. Crops are resized to $224\times224$ and normalised with ImageNet statistics. The frozen embeddings are classified by a lightweight five-layer feed-forward network with ReLU activations and dropout ($0.2$--$0.3$) for binary detection (bona fide vs.\ attack). All FLiD classification heads are trained for up to $100$ epochs using Adam (learning rate $10^{-3}$, weight decay $10^{-4}$, batch size $32$) and \texttt{BCEWithLogitsLoss} with \texttt{pos\_weight} rebalancing. Early stopping (patience 15) and the ReduceLROnPlateau schedule are driven by an inner validation split---15\% of the training fold, partitioned by document identity, so the held-out test fold is never used for model selection. The same protocol is applied to the baseline. Experiments run in PyTorch~2.10.0 on an NVIDIA H100 NVL GPU.

\paragraph{Baseline.}
Our primary baseline reimplements González and Tapia~\cite{pad2025pattern}:
a MobileNetV2 trained from scratch on full-document images at $448\times448$ with a two-class
softmax head (Adam, learning rate $10^{-5}$). For combined attacks it uses
the same score-level fusion as FLiD,
$\min(P_{\text{face}}(\text{real}),P_{\text{text}}(\text{real}))$. All
baseline models use identical splits and metrics for a fair comparison.

\textit{Computational comparison.}
The full-document baseline contains $2.55$M trainable parameters and requires $2{,}503$M FLOPs per pass. In contrast, each FLiD detector trains only a $191$K-parameter head and requires $119$M FLOPs per field, corresponding to $13\times$ fewer trainable parameters and $21\times$ fewer FLOPs per field. Even for a full multi-field document, FLiD remains substantially more efficient because the text head is shared across sub-fields.

\paragraph{Evaluation metrics.}
We adopt the ISO/IEC 30107-3~\cite{biometrics2017information,
biometrics2016iso} PAD protocol, consistent with~\cite{pad2025pattern}. We
report APCER (attack samples accepted as bona fide) and BPCER (genuine
samples rejected as attacks), and derive the Equal Error Rate (EER) at
their crossover. To characterise the security--usability trade-off at a
fixed operating point, we additionally report BPCER$_{10}$ (BPCER at an
APCER of $10\%$). We complement these threshold-based metrics with Accuracy and ROC AUC, which summarise overall decision and ranking performance.

\paragraph{Dataset.}
We evaluate on FantasyID~\cite{korshunov2025fantasyid}, a publicly available benchmark of $362$ synthetic ``fantasy'' identity documents generated without any real personal data, making it suitable for privacy-safe forgery-detection research and adopted as the benchmark for the IEEE/CVF ICCV DeepID Challenge. Each document contains a facial photograph and structured biographic text fields, with bona-fide samples and three localized manipulation types: face replacement, text-field editing, and their combination.

\subsection{Detection Performance}
Table~\ref{tab:detection} reports detection performance across threshold-based (EER, BPCER$_{10}$) and ranking/decision (AUC, Accuracy) metrics. The from-scratch full-document baseline collapses to near-random discrimination (EER $\approx48$--$53\%$, AUC $0.47$--$0.53$) across all scenarios, whereas FLiD retains strong discrimination throughout, lowering EER by $28.0$--$29.1$~pp and raising AUC to $0.83$--$0.93$.
Text attacks show the largest gains and are FLiD's strongest scenario, with the baseline near chance and FLiD cutting BPCER$_{10}$ by $63$~pp, indicating that localised textual edits leave strongly detectable traces even under automatic localisation. Face attacks show a smaller but still substantial margin over the baseline. Under the cross-attack combined setting, FLiD remains fully competitive with single-field detection, since two simultaneously forged fields give the most-suspicious-field rule two independent chances to flag the document.

FLiD's consistent margin over this baseline, across threshold, ranking, and decision metrics, and at every operating point, demonstrates that field localisation captures genuine, transferable tampering cues rather than dataset-specific artefacts.

\begin{table*}[t]
\centering
\caption{Detection performance under 5-fold cross-validation: Baseline vs.\ the proposed FLiD
with automatic YOLO11 field detection. EER and BPCER$_{10}$ follow ISO/IEC
30107-3; AUC and Accuracy are reported as mean\,$\pm$\,std. For EER and
BPCER$_{10}$, $\Delta$ is the absolute change in percentage points
(negative $=$ error reduction). Best per scenario in \textbf{bold}.}
\label{tab:detection}
\begin{tabular}{l ccc ccc ccc}
\toprule
& \multicolumn{3}{c}{\textbf{Face Attack}}
& \multicolumn{3}{c}{\textbf{Text Attack}}
& \multicolumn{3}{c}{\textbf{Face + Text Attack}} \\
\cmidrule(lr){2-4}\cmidrule(lr){5-7}\cmidrule(lr){8-10}
Metric & Base & FLiD & $\Delta$ & Base & FLiD & $\Delta$ & Base & FLiD & $\Delta$ \\
\midrule
EER (\%) $\downarrow$
 & 53.36 & \textbf{24.68} & $-28.68$
 & 47.53 & \textbf{18.46} & $-29.07$
 & 51.18 & \textbf{23.20} & $-27.98$ \\
BPCER$_{10}$ (\%) $\downarrow$
 & 93.00 & \textbf{43.00} & $-50.00$
 & 90.03 & \textbf{27.03} & $-63.00$
 & 89.15 & \textbf{46.85} & $-42.30$ \\
\midrule
AUC $\uparrow$
 & 0.469\tiny$\pm$.107 & \textbf{0.834}\tiny$\pm$.077 & --
 & 0.532\tiny$\pm$.032 & \textbf{0.926}\tiny$\pm$.003 & --
 & 0.507\tiny$\pm$.063 & \textbf{0.837}\tiny$\pm$.055 & -- \\
Accuracy (\%) $\uparrow$
 & 50.37\tiny$\pm$9.5 & \textbf{75.20}\tiny$\pm$7.8 & --
 & 52.42\tiny$\pm$3.4 & \textbf{81.54}\tiny$\pm$0.3 & --
 & 48.82\tiny$\pm$6.3 & \textbf{76.80}\tiny$\pm$6.5 & -- \\
\bottomrule
\end{tabular}
\end{table*}

\subsection{Score Separability and Ranking Reliability}
\begin{figure*}[htbp]
    \centering
    \begin{subfigure}[t]{0.35\linewidth}
        \centering
        \includegraphics[width=\linewidth]{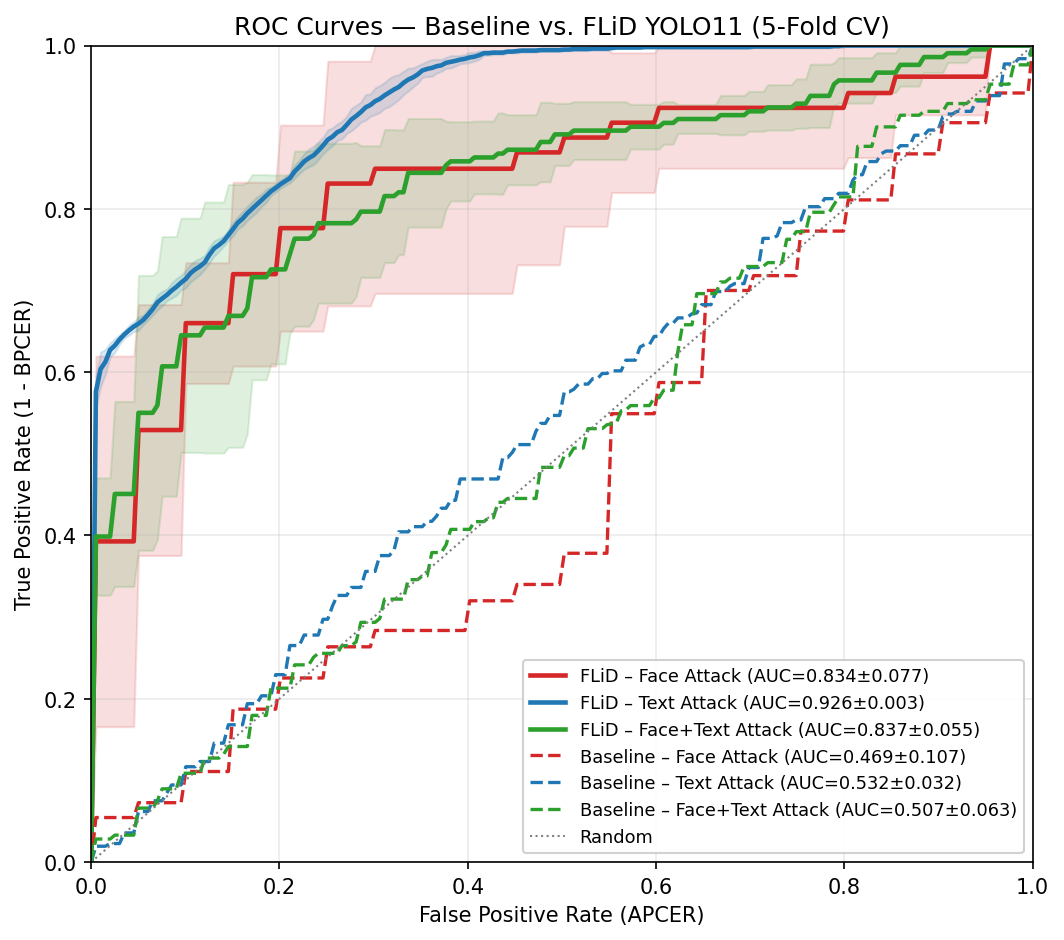}
        \caption{ROC curves}
        \label{fig:roc_sub}
    \end{subfigure}
    \hfill
    \begin{subfigure}[t]{0.6\linewidth}
        \centering
        \includegraphics[width=\linewidth]{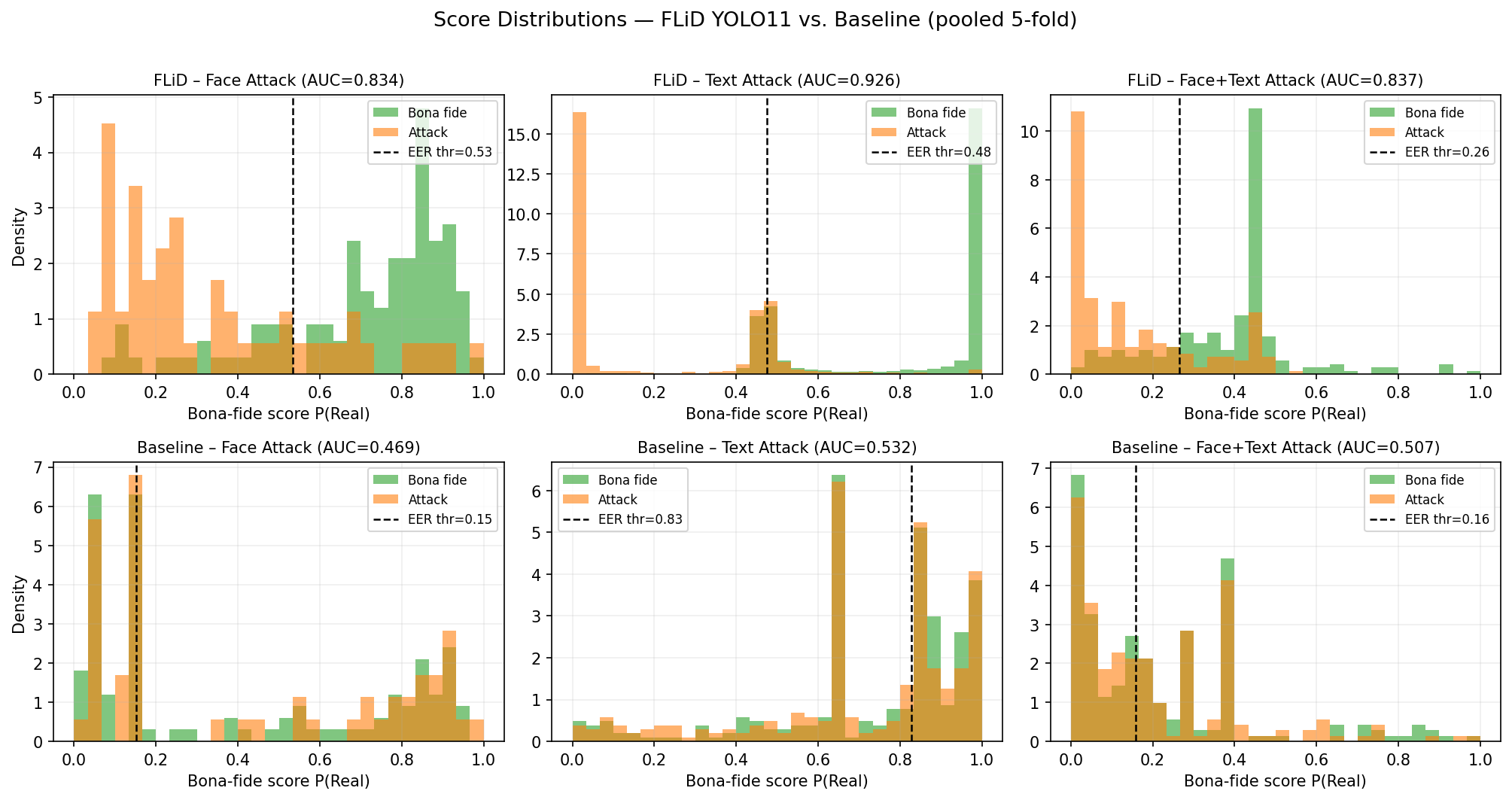}
        \caption{Score distributions}
        \label{fig:score_sub}
    \end{subfigure}
    \caption{Comparison of FLiD and the baseline across face, text, and combined-field attack scenarios. (a) ROC curves illustrate ranking performance. (b) Score distributions show class separability and overlap.}
    \label{fig:roc_score_combined}
\end{figure*}

Figure~\ref{fig:roc_sub} and Figure~\ref{fig:score_sub} together show that FLiD improves both ranking quality and score separability over the full-document baseline. FLiD's ROC curves remain well above the diagonal across all three attack scenarios, whereas the baseline stays close to chance. The text attack is the most separable case, with the highest and most stable AUC ($0.926\pm0.003$) and a near-bimodal score distribution yielding the lowest EER ($18.5\%$); face and combined attacks remain comparable to each other in both AUC ($0.834$ vs.\ $0.837$) and EER ($24.7\%$ vs.\ $23.2\%$), confirming that the most-suspicious-field rule preserves single-field performance under fusion. In contrast, baseline scores overlap heavily across all scenarios, explaining its near-chance AUC and confirming that field localisation improves both ranking quality and decision confidence over full-document modelling.

\subsection{Backbone and Crop-Strategy Ablation}
\label{subsec:ablation}
 We evaluated MobileNetV3-Small, EfficientNet-B0, and ResNet50 as feature backbones, as shown in Table~\ref{tab:backbone_crop_ablation}. Under the deployable YOLO11 setting, all three perform comparably on single-field attacks (EfficientNet-B0 best on face: AUC $0.85$, EER $21.95\%$; all $\approx0.93$ AUC on text), but this does not hold for combined attacks: MobileNetV3-Small reaches AUC $0.84$ (EER $23.20\%$), versus $0.76$ (EfficientNet-B0) and $0.67$ (ResNet50). Larger backbones thus give only marginal single-field gains while reducing robustness once face and text decisions are composed. We retain MobileNetV3-Small as it remains competitive on face and text, performs best on the combined scenario, and is the most efficient to deploy.

\begin{table}[htbp]
\centering
\scriptsize
\caption{Backbone and crop-strategy ablation. GT denotes manually annotated field crops, YOLO11 the automatic deployable setting, and Coarse a broad text-region crop. Both denotes combined face--text attacks; EER is reported in \%. Coarse and GT use the same face crop. }
\setlength{\tabcolsep}{2.3pt}
\begin{tabular}{ll|cc|cc|cc}
\hline
\multirow{2}{*}{\textbf{Backbone}} & \multirow{2}{*}{\textbf{Crop}}
  & \multicolumn{2}{c|}{\textbf{Face Attack}}
  & \multicolumn{2}{c|}{\textbf{Text Attack}}
  & \multicolumn{2}{c}{\textbf{Both Attack}} \\
 &  & AUC & EER & AUC & EER & AUC & EER \\
\hline
\multirow{3}{*}{MobileNetV3-Small}
 & GT           & 0.86 & 22.45 & 1.00 &  0.98 & 0.89 & 19.89 \\
 & YOLO11       & 0.83 & 24.68 & \textbf{0.93} & 18.46 & \textbf{0.84} & \textbf{23.20} \\
 & Coarse       & 0.86 & 22.45 & 0.72 & 35.80 & 0.79 & 27.00 \\
\hline
\multirow{3}{*}{EfficientNet-B0}
 & GT           & 0.84 & 22.86 & 1.00 &  0.60 & 0.77 & 31.26 \\
 & YOLO11       & \textbf{0.85} & \textbf{21.95} & \textbf{0.93} & \textbf{17.64} & 0.76 & 29.39 \\
 & Coarse       & 0.84 & 22.86 & 0.75 & 33.88 & 0.72 & 35.06 \\
\hline
\multirow{3}{*}{ResNet50}
 & GT           & 0.80 & 27.59 & 1.00 &  1.03 & 0.73 & 32.96 \\
 & YOLO11       & 0.80 & 28.09 & 0.93 & 18.08 & 0.67 & 39.57 \\
 & Coarse       & 0.80 & 27.59 & 0.66 & 39.36 & 0.62 & 41.95 \\
\hline
\end{tabular}
\label{tab:backbone_crop_ablation}
\end{table}

Crop strategy has the largest effect on text attacks: AUC falls from $1.00$ (GT) to $0.93$ (YOLO11) to $0.66$--$0.75$ (Coarse), showing text-attack performance is primarily limited by localisation precision rather than classification. Face-attack AUC is nearly unaffected by crop source (GT vs.\ YOLO11 differs by $\leq0.03$), indicating YOLO11 already localises faces reliably. Combined-attack performance follows the same ordering (GT $0.89>$ YOLO11 $0.84>$ Coarse $0.79$), confirming that more precise field localisation benefits combined-attack detection as well.

\textbf{Comparison with General-Purpose Forgery Detectors.} We compare FLiD with three general-purpose image-forgery detectors, TruFor, MMFusion, and UniVAD, under the same protocol using their pretrained models. FLiD achieves the highest Accuracy and Precision in every scenario, outperforming the strongest baseline, TruFor, by $+10.7$~pp (face), $+4.7$~pp (text), and $+15.4$~pp (combined) in Accuracy, with even larger Precision gains of $+11.9$, $+13.6$, and $+18.9$~pp respectively. MMFusion and UniVAD remain close to chance across most settings, consistent with the original FantasyID evaluation~\cite{korshunov2025fantasyid}. These results show that detectors designed for natural-image manipulations do not capture the localized, semantically constrained edits found in identity documents, and confirm FLiD's field-localised design as a more reliable and lightweight alternative for KYC-oriented deployment.

\subsection{Limitations and Future Directions}

The results of FLiD suggest several directions for extending field-localised identity-document forgery detection. FLiD currently focuses on visual appearance cues within semantic fields, which is effective for detecting face and text manipulations involving font, blending, texture, or resolution inconsistencies. As generative editing improves, future systems could strengthen this representation with OCR-derived semantic signals for visual--linguistic consistency checks between document images and extracted text.

Future work could also move beyond field-level decisions by producing patch-level tampering maps that indicate where an edit occurs within a field. Combined with explainability methods such as Grad-CAM or attention-based saliency, such evidence could support human reviewers in operational KYC workflows. Finally, although FantasyID provides a diverse and privacy-preserving benchmark, broader cross-dataset evaluation across document formats, capture conditions, and manipulation methods would further establish the generality of field-localised detection in real-world deployment.

\section{Conclusion}
\label{sec:conclusion}
We introduced FLiD, a lightweight field-localised framework for detecting face, text, and combined forgeries in structured digital identity documents. Unlike full-document approaches, FLiD explicitly models identity documents as a composition of semantic fields, localising face and text regions with YOLO11, encoding them with a frozen MobileNetV3-Small backbone, and classifying field-level manipulations using a compact $191$K-parameter head. This design substantially outperforms a from-scratch full-document baseline and general-purpose natural-image forgery detectors, at a fraction of the trainable parameters and FLOPs. These results demonstrate that field-level analysis is a more effective and deployment-oriented strategy for identity-document forgery detection, particularly in data- and resource-constrained KYC settings.

\section*{ CRediT authorship contribution statement}
\textbf{Abhishek Kumar:} Writing – review \& editing, Writing – original draft, Validation, Methodology, Data curation, Conceptualization. \textbf{Riya Tapwal}: Writing – review \& editing, Methodology, Data curation. \textbf{Carsten Maple}: Validation, Supervision, Methodology. \textbf{Mark Hooper}: Writing – review \& editing, Validation, Supervision.

\section*{ Declaration of generative AI and AI-assisted technologies in the manuscript preparation process}
The author(s) used ChatGpt to improve readability and grammar. The author(s) are fully responsible for the publication's content after using this tool/service, including reviewing and editing it as required.

\section*{Declaration of competing interest}
The authors declare that they have no known competing financial interests or personal relationships that could have appeared to influence the work reported in this paper.

\section*{Data availability}

The data is provided in our Github repository\cite{flid_results_repository}.

\bibliographystyle{cas-model2-names}
\bibliography{cas-refs}

\appendix

\end{document}